\documentclass[onecolumn]{IEEEconf} 
\usepackage{nips12submit_e,times}
\usepackage[latin1]{inputenc}
\usepackage{amsmath}
\usepackage{amsfonts}
\usepackage{amssymb}
\usepackage{color}
\usepackage{graphicx}
\usepackage{bm}
\usepackage{algorithm,algorithmic}
\usepackage{multicol,multirow}
\usepackage{subfigure}
\usepackage{enumerate}

\usepackage{soul}
  \usepackage{color}
\definecolor{lightblue}{rgb}{.5,.85,1}
\definecolor{lightred}{rgb}{1,.4,.3}
\definecolor{lightorange}{rgb}{1,.7,.4}
\definecolor{yellow}{rgb}{1,1,0}

\providecommand{\rem}[1]{}%
\providecommand{\com}[1]{}

\DeclareMathOperator{\EV}{\mathrm{E}}  
\newtheorem{thm}{Theorem}[section]
\newtheorem{prop}{Proposition}[section]
\newtheorem{lmm}{Lemma}[section]
\newtheorem{prf}{Proof}[section]

\newtheorem{defi}{Definition}[section]

\title{Information Theoretic Learning with Infinitely Divisible Kernels}

\author{
Luis G. {Sanchez Giraldo}
\\
 Dept. of Electrical and Computer Engineering\\
 University of Florida\\
 Gainesville, Florida, USA\\
\texttt{sanchez@cnel.ufl.edu} \\
\And
Jose C. Principe \\
 Dept. of Electrical and Computer Engineering\\
 University of Florida\\
 Gainesville, Florida, USA\\
\texttt{principe@cnel.ufl.edu} \\
}

\nipsfinalcopy 

\begin{document}
\maketitle

\begin{abstract}
 In this paper, we develop a framework for information theoretic learning based on infinitely divisible matrices. We formulate an entropy-like functional on positive definite matrices based on Renyi's axiomatic definition of entropy and examine some key properties of this  functional that lead to the concept of infinite divisibility. The proposed formulation avoids the plug in estimation of density and brings along the representation power of reproducing kernel Hilbert spaces. As an application example, we derive a supervised metric learning algorithm using a matrix based analogue to conditional entropy achieving results comparable with the state of the art. 
\end{abstract}

\section{Introduction}\label{sec:Introduction}
Information theoretic quantities are descriptors of the distributions of the data that go beyond second-order statistics. The expressive richness of quantities such entropy or mutual information has been shown to be very useful for machine learning problems where optimality based on linear and Gaussian assumptions no longer holds. Nevertheless, operational quantities in information theory are based on the probability laws underlying the data generation process, which are rarely known in the statistical learning setting where the only information available comes from the sample $\{z_i\}_{i=1}^n$. Therefore, the use of information theoretic quantities as descriptors of data requires the development of suitable estimators. In \cite{JPrincipe}, the use of Renyi's definition of entropy along with Parzen density estimation is proposed as the main tool for information theoretic learning (ITL). The optimality criteria is expressed in terms of quantities such as Renyi's entropy, divergences based on the Cauchy-Schwarz inequality, quadratic mutual information, among others. Part of the research effort in this context has pointed out connections to reproducing kernel Hilbert spaces \cite{JWXu08}. Here, we show that these connections are not only valuable from a theoretical point of view, but they can also be exploited to derive novel information theoretic quantities with suitable estimators from data.

Positive definite kernels have been employed in machine learning as a representational tool allowing algorithms that are based on inner products to be expressed in a rather generic way (the so called \emph{``kernel trick"}). Algorithms that exploit this property are commonly known as kernel methods.  Let   $\mathcal{X}$ be a nonempty set. A function $\kappa: \mathcal{X} \times \mathcal{X} \mapsto \mathbb{R}$ is called a positive definite kernel if for any finite set $\{x_i\}_{i=1}^{N} \subseteq \mathcal{X}$ and any set of coefficients $\{\alpha_i\}_{i=1}^{N} \subset \mathbb{R}$, it follows that $\sum_{i,j}\alpha_i\alpha_j\kappa(x_i,x_j) \geq 0$, if at least one $i$, $\alpha_i \neq 0$. In this case, there exist an implicit mapping $\phi: \mathcal{X} \mapsto \mathcal{H}$ that maps any element $x \in \mathcal{X}$ to an element $\phi(x)$ in a Hilbert space $\mathcal{H}$, such that $\kappa(x,y) = \langle\phi(x),\phi(y) \rangle$. The above map provides an implicit representation of the objects of interest that belong to the set $\mathcal{X}$. The generality of this representation has been exploited in many practical applications, even for data that do not come in standard vector representation $\mathbb{R}^d$ \cite{JShaweTaylor}. This is possible  as long as a kernel function is available.

More recently, it has been noticed that kernel induced maps are also useful beyond the above kernel trick in a rather interesting fashion. Namely, kernels can be utilized to compute higher-order statistics of the data in a nonparametric setting. Some examples exploring this idea are: kernel independent component analysis \cite{FBach02}, the work on measures of dependence and independence using Hilbert-Schmidt norms \cite{AGretton05}, and the quadratic measures of independence proposed in \cite{SSeth11}. It is not surprising, yet important to mention, that a similar observation have also been reached from the work on ITL since one of the original motivations in using information theoretic quantities is to go beyond second order statistics. The work we introduce in this paper goes along these lines. The twist is that rather than defining an estimator of a conventional information theoretic quantity such as Shannon entropy, we propose a quantity build from the data that satisfies similar axiomatic properties to those of well establish definitions such as Renyi's definition of entropy
 
The main contribution of this work is to show that the Gram matrix obtained from evaluating a positive definite kernel on samples can be used to define a quantity based on the data with properties similar to those of an entropy without assuming that the probability density is being estimated. Therefore, we look at the axiomatic treatment of entropy and adapt it to the Gram matrices describing the data. In this sense, we think about entropy as a measure inversely related to the amount of statistical regularities (structure) directly from the data that can be applied as the optimality criterion in a learning algorithm. 
As an application example, we derive supervised metric learning algorithm that uses conditional entropy as the cost function. This is the second contribution of this paper, and the empirical results show that the proposed method is competitive with current approaches.\\
The main body of the paper is organized in two parts. First, we introduce the proposed matrix-based entropy measure using the spectral theorem along with a set of axiomatic properties that our quantity must satisfy. Then, the notion of joint entropy is developed based on Hadamard products. We look at some basic inequalities of information and how they translate to the setting of positive definite matrices, which finally allow us to define an analogue to conditional entropies. In the development of these ideas, we find that the concept of infinitely divisible kernels arises and become key to our purposes. We revisit some of the theory on infinitely divisible matrices, to show how it links to the the proposed information theoretic framework. 
In the last part, we introduce an information theoretic supervised metric learning algorithm. We show how the proposed analogue to conditional entropy is a suitable cost function leading naturally to a gradient descent procedure.
Finally, we provide some conclusions and future directions.

\section{Positive Definite Matrices, and Renyi's Entropy Axioms}
Let us start with an informal observation that motivated our matrix based entropy. In \cite{JPrincipe}, the use of Renyi's entropy is proposed as an alternative to the more commonly adopted definition of entropy given by Shannon. In particular, it was found that Renyi's second-order entropy provides an amenable quantity for practical purposes. An empirical plug in estimator of Renyi's second-order entropy based on the Parzen density estimator $\hat{f}(x) = \frac{1}{n}\sum_{i=1}^n \kappa(x_i,x),$ can be obtained as follows: 
\begin{equation}\label{eq:Renyi_entropy_estimator}
-\log{\frac{1}{n^2}\sum\limits_{i,j=1}^n h(x_i,x_j)},
\end{equation}
where $h(x,y) = \int_{\mathcal{X}}\kappa_{\sigma}(x,z)\kappa_{\sigma}(y,z)\mathrm{d}z$. Note that since $h$ is a positive definite kernel, there exists a mapping $\phi$ to a RKHS such that $h(x,y) = \langle \phi(x),\phi(y)\rangle$; and the argument of the $\log$ in \eqref{eq:Renyi_entropy_estimator}, called the \emph{information potential}, can be interpreted  in this space as a norm: 
\begin{equation}\label{eq:information_potential_RKHS_estimator}
\left\langle\frac{1}{n}\sum_{i=1}^n\phi(x_i), \frac{1}{n}\sum_{i=1}^n\phi(x_i)\right\rangle = \left\Vert \frac{1}{n}\sum_{i=1}^n\phi(x_i)\right\Vert^2,
\end{equation}
with the limiting case given by $\Vert \EV[\phi(X)]\Vert^2$. Thus, we can think of this estimator as an statistic computed on the representation  space provided by the positive definite kernel $h$. Now, let us look at the case where $\kappa_{\sigma}$ is the Gaussian kernel; if we construct the Gram matrix $\mathbf{K}$ with elements $K_{ij} = \kappa_{2\sigma}(x_i,x_j)$, it is easy to verify that the estimator of Renyi's second-order entropy based on \eqref{eq:Renyi_entropy_estimator} corresponds to:
\begin{equation}\label{eq:trace_second_renyi_entropy}
\hat{H}_2(X) = -\log{\left(\frac{1}{n^2}\mathrm{tr}{\left(\mathbf{K}\mathbf{K}\right)}\right)} + C(\sigma).
\end{equation}
where $C(\sigma)$ takes care of the normalization factor of the Parzen window. As we can see, the information potential estimator can be related to the norm of the Gram matrix $\mathbf{K}$ defined as $\Vert\mathbf{K}\Vert^2 = \mathrm{tr}{\left(\mathbf{K}\mathbf{K}\right)}$. From the above informal argument two important questions arise. First, it seems natural to ask whether other functionals on Gram matrices allow information theoretic interpretations that can be further utilized as objective functions in ITL. Secondly, even though $h$ was originally derived from a convolution of Parzen windows, was there anything about the implicit representation that allows to interpret \eqref{eq:information_potential_RKHS_estimator} in information theoretic terms?   
\subsection{Renyi's Axioms for Gram matrices}
Real Hermitian matrices are considered generalizations of real numbers. It is possible to define a partial ordering on this set by using positive definite matrices, which are a generalization of the positive real numbers. Let $M_n$ be the set of all $n\times n$ real matrices; for two Hermitian matrices $A,B \in M_n$, we say $A \succcurlyeq B$ if $A - B$ is positive definite. Likewise, $A \succ B$ means that $A - B$ is strictly positive definite.

The following spectral decomposition theorem \cite{RHorn_Topics_in_Matrix_Analysis} relates to the functional calculus on matrices and provides a reasonable way to extend continuous scalar-valued functions to Hermitian matrices.  
\begin{thm}\label{thm:functional_calculus_psd_matrix} Let $D\subset \mathbb{C}$ be a given set and let $\mathcal{N}_n(D) := \{A \in M_n:\:A \textrm{ is normal and }\: \sigma(A)\subset D\},$ where $\sigma(A)$ denotes the spectrum of $A$. If $f(t)$ is a continuous scalar-valued function on $D$, then the primary matrix function
\begin{equation}\label{eq:primary_matrix_function}
f(A) = U \left(\begin{array}{ccc}
f(\lambda_1) & \cdots & 0 \\
\vdots & \ddots & \vdots \\
0 & \cdots & f(\lambda_n) \\
\end{array}\right)U^*
\end{equation}
is continuous on $\mathcal{N}_n(D)$, where $A = U \Lambda U^*$, $\Lambda = diag(\lambda_1,\dots,\lambda_n)$, and $U \in M_n$ is unitary.
\end{thm}
Equipped with the above result, we can define matrix functions such as $f(A) = A^r$ for $r \in \mathbb{R}^{+}$, which will be used in defining the following matrix-based analogue to Renyi's $\alpha$-entropy. The functional will then be applied to  Gram matrices constructed by pairwise evaluation of a positive definite kernel on the data samples.    

Consider the set $\Delta_n^{+}$ of positive definite matrices $A \in M_n$ for which $\mathrm{tr}{(A)} \leq 1$. It is clear that this set is closed under finite convex combinations.
\begin{prop}\label{prop:matrix_entropy} Let $A \in \Delta_n^{+}$ and $B \in \Delta_n^{+}$ and also $\mathrm{tr}{(A)} = \mathrm{tr}{(B)} = 1$. The functional
\begin{equation}\label{eq:renyi_matrix_entropy}
S_{\alpha}(A) = \frac{1}{1-\alpha}\log_{2}{\left[\mathrm{tr}{(A^{\alpha})}\right]},
\end{equation}
satisfies the following set of conditions:
\begin{enumerate}[(i)]
\item \label{property:matrix_symmetry} $S_{\alpha}(PAP^{*}) = S_{\alpha}(A)$ for any orthonormal matrix $P \in M_n$  
\item \label{property:matrix_continuity} $S_{\alpha}(pA)$ is a continuous function for $0 < p \leq 1$.
\item \label{property:matrix_max} $S_{\alpha}(\frac{1}{n}I) = \log_2{n}$.
\item \label{property:matrix_joint} $S_{\alpha}(A \otimes B) = S_{\alpha}(A) + S_{\alpha}(B)$.
\item \label{property:matrix_generalized_mean_value} If $AB = BA = \mathbf{0}$; then for the strictly monotonic and continuous function $g(x) = 2^{(\alpha-1)x}$ for $\alpha \neq 1$ and $\alpha > 0$, we have that:
\begin{equation}\label{eq:matrix_entropy_mean_value}
\begin{split}
S_\alpha(tA+(1-t)B) =  g^{-1}\left( tg(S_\alpha(A)) + (1-t)g(S_\alpha(B))\right).
\end{split}
\end{equation} 
\end{enumerate}
\begin{prf} The proof of (\ref{property:matrix_symmetry}) easily follows from Theorem \ref{thm:functional_calculus_psd_matrix}. Take $A = U \Lambda U^*$ now $PU$ is also a unitary matrix and thus $f(A)$ = $f(PAP^*)$ the trace functional is invariant under unitary transformations. For (\ref{property:matrix_continuity}), the proof reduces to the continuity of $\frac{1}{1-\alpha}\log_2(p)^\alpha$. For  (\ref{property:matrix_max}), a simple calculation yields $\mathrm{tr}{A^{\alpha}} = \left(\frac{1}{n}\right)^{\alpha-1}$. Now,  for property (\ref{property:matrix_joint}), notice that if $\mathrm{tr}{A} = \mathrm{tr}{B} = 1$, then,  $\mathrm{tr}{(A \otimes B)} = 1$. Since $A = U \Lambda U^*$ and $B = V \Gamma V^*$ we can write $A \otimes B = (U \otimes V)(\Lambda \otimes \Gamma)(U \otimes V)^*$, from which $\mathrm{tr}{(A \otimes B)^{\alpha}} = \mathrm{tr}{(\Lambda \otimes \Gamma)^{\alpha}} = \mathrm{tr}{(\Lambda^{\alpha})}\mathrm{tr}{(\Gamma^{\alpha})}$ and thus (\ref{property:matrix_joint}) is proved. Finally, (\ref{property:matrix_generalized_mean_value}) notice that for any integer power $k$ of $tA+(1-t)B$ we have: $(tA+(1-t)B)^k = (tA)^k + ((1-t)B)^k$ since $AB = BA = \mathbf{0}$. Under extra conditions such as $f(0) = 0$ the argument in the proof of Theorem \ref{thm:functional_calculus_psd_matrix} can be extended to this case. Since the eigen-spaces for the non-null eigenvalues of $A$ and $B$ are orthogonal we can simultaneously diagonalize $A$ and $B$ with the orthonormal matrix $U$, that is $A = U\Lambda U^*$ and $B = U\Gamma U^*$ where $\Lambda$ and $\Gamma$ are diagonal matrices containing the eigenvalues of $A$ and $B$ respectively. Since $AB = BA = \mathbf{0}$, then $\Lambda \Gamma = \mathbf{0}$. Under the extra condition $f(0) = 0$, we have that $f(tA+(1-t)B) = f(tA) + f((1-t)B)$ yielding the desired result for (\ref{property:matrix_generalized_mean_value}).
\end{prf}
\end{prop} 
Notice also that if the rank of $A$,  $\rho(A) = 1$, the entropy $S_{\alpha}(A) = 0$ for any $\alpha \neq 0$. \\
It is also true that,
\begin{equation}
S_{\alpha}(A) \leq S_{\alpha}(\frac{1}{n}I)=\log_2{n}.
\end{equation}
As we can see \eqref{eq:renyi_matrix_entropy} satisfies some properties attributed to entropy. Nevertheless, such a characterization may not fully endow all unit-trace positive definite matrices with an information theoretic interpretation. Which descriptors are suitable in representing joint-spaces? What properties should be satisfied by the matrices in order to be applied to concepts that link them to random variables such as conditioning? In what follows, we address these points by developing notions of joint entropy and conditional entropy, for which, additional properties must be fulfilled. Recall that the notions of joint and conditional entropy are not only important for the above reasons, but they also provide the means to propose objective functions for learning that are based on information theoretic quantities.
\subsection{Hadamard Products and the Notion of Joint Entropy}
Positive kernels are also useful in integrating multiple modalities. Using the the product kernel, we can readily define the notion of joint-entropy. Consider a sequence of sample pairs $\{(x_i,y_i)\}_{i=1}^N$ where $x_i \in \mathcal{X}$ and $y_i \in \mathcal{Y}$. Assume, we have a positive definite kernels $\kappa_1$ defined on $\mathcal{X}\times\mathcal{X}$ and $\kappa_2$ defined on $\mathcal{X}\times\mathcal{X}$. The product kernel $\kappa((x_i,y_i),(x_j,y_j)) =  \kappa_1(x_i,x_j) \kappa(y_i,y_j)$ is a positive definite kernel on $(\mathcal{X}\times \mathcal{Y})\times(\mathcal{X}\times \mathcal{Y})$. As we can see the Hadamard product arises as a joint representation in a our matrix based entropy. Consider two matrices $A$ and $B$ in $\Delta_n$ with unit trace, for which there exists some relation between the elements $A_{ij}$ and $B_{ij}$ for all $i$ and $j$. 
The joint entropy can be defined as:
\begin{equation}\label{eq:hadamard_joint_entropy}
S_{\alpha}\left(\frac{A \circ B}{\mathrm{tr}(A \circ B)}\right)
\end{equation}
It is important then to verify that the definition of joint entropy \eqref{eq:hadamard_joint_entropy} satisfies a basic intuition about uncertainty. The joint entropy should never be smaller than any of the individual entropies of the variables that conform it.  The following proposition verifies this intuition for a subset of the unit trace, positive definite matrices.
\begin{prop} Let $A$ and $B$ be two $n \times n$ positive definite matrices with trace $1$ with nonnegative entries, and $A_{ii} = \frac{1}{n}$ for $i=1,2,\dots,n$. Then, the following inequality holds:
 \begin{equation}\label{eq:hadamard_conditional_entropy_inequality} 
 S_{\alpha}\left(\frac{A \circ B}{\mathrm{tr}(A \circ B)}\right) \geq S_{\alpha}(B),
\end{equation}
\end{prop}

\subsection{Conditional Entropy as a Difference Between Entropies }
The conditional entropy of $X$ given $Y$, which can be understood as the uncertainty about $X$ that remains after knowing the joint distribution of $X$ and $Y$, can be obtained from a difference between two entropies. In the Shannon's definition of conditional entropy,  $H(X|Y)$ can be expressed as $H(X|Y) = H(X,Y)-H(Y)$. The properties of this definition has been recently studied in the case of Renyi's entropies \cite{ATeixeira12} and in the matrix case, this definition yields:
\begin{equation}\label{eq:conditonal_entropy_gap}
S_{\alpha}(A|B) =  S_{\alpha}\left(\frac{A \circ B}{\mathrm{tr}{(A\circ B)}}\right) - S_{\alpha}(B),
\end{equation}
for positive semidefinite matrices $A$ and $B$ with nonnegative entries and unit trace, such that $A_{ii} = \frac{1}{n}$ for all $i=1,\dots,n$. The above quantity is nonnegative and upper bounded by $S_{\alpha}(A)$. Certainly, normalization is an important property of the matrices involved in the above results. If $A$ and $B$ are normalized to have unit trace, then for $r \in [0,1]$ it is true that the Hadamard product of 
\begin{equation}\label{eq:hadamard_geometric_average}
A^{\circ r}\circ B^{\circ(1-r)},
\end{equation}
is also normalized. However, it is not always true that the resulting matrix \eqref{eq:hadamard_geometric_average} is positive definite. This product can be thought as a weighted geometric average for which the resulting matrix will give more emphasis to either one of the matrices. However, if $A$ and $B$ satisfy a property called infinitely divisibility, the product is guaranteed to be positive definite \footnote{By this, we also mean positive semidefinite}.  

\section{Infinitely Divisible Functions}
The theory of infinitely divisible developed below is not new, but it is included because it provides a basic understanding about the role of infinitely divisible kernels in computing the above information theoretic quantities from data. To avoid confusion, let us describe the key points to bear in mind before we move to the mathematical description. Infinitely divisible kernels and negative definite functions are tied together trough the exponential a logarithm functions. Both functions provide Hilbert space representations of the data. We can think of the RKHS of the infinitely divisible kernel as a representation to compute the higher order descriptors of the data. On the other hand, the Hilbertian metric can be the representation space for which we want to compute the high order statistics. Normalization, as we show below is not only important in satisfying the conditions for the information theoretic quantities already defined, but it also shows that many possible representational choices are equivalent.

\subsection{Negative Definite Functions and Hilbertian Metrics}
Let $\mathcal{M} = \left( \mathcal{X},d \right)$ be a separable metric space. A necessary and sufficient condition for $\mathcal{M}$ to be embeddable in a Hilbert space $\mathcal{H}$ is that for any set $\{ x_i \} \subset \mathcal{X}$ of $n+1$ points, $
\sum_{i,j =1}^n\alpha_i\alpha_j\left(d^2(x_0,x_i) + d^2(x_0,x_j) - d^2(x_i,x_j) \right) \geq 0,
$ for any $\bm{\alpha} \in \mathbb{R}^n$. This condition is equivalent to $\sum_{i,j = 0}^n\alpha_i\alpha_j d^2(x_i,x_j)  \leq 0,$ for any $\bm{\alpha} \in \mathbb{R}^{n+1}$, such that $\sum_{i=0}^n\alpha_i = 0$. This condition is known as negative definiteness. Interestingly, the above condition implies that $\exp(-r d^2(x_i,x_j))$ is positive definite in $\mathcal{X}$ for all $r>0$ \cite{ISchoenberg38}. Indeed, matrices derived from functions satisfying the above property conform a special class of matrices know as infinitely divisible.
\subsection{Infinitely Divisible Matrices}
According to the Schur product theorem $A \succcurlyeq 0$ implies $A^{\circ n} = A \circ A \circ \cdots \circ A \succcurlyeq 0$ for any positive integer $n$. Does the above hold if we to take fractional powers of $A$? In other words,is the matrix $A^{\circ \frac{1}{m}} \succcurlyeq 0$ for any positive integer $m$? This question leads to the concept of infinitely divisible matrices \cite{RBhatia06,RHorn69}. A nonnegative matrix $A$ is said to be infinitely divisible if $A^{\circ r} \succcurlyeq 0$ for every nonnegative $r$.
Infinitely divisible matrices are intimately related to negative definiteness as we can see from the following proposition
\begin{prop}\label{prop:infinite_divisibility_negative_definiteness} If $A$ is infinitely divisible, then the matrix $B_{ij} =-\log{A_{ij}}$ is negative definite 
\end{prop} 
From this fact it is possible to relate infinitely divisible matrices with isometric embeddings into Hilbert spaces. If we construct the matrix
\begin{equation}\label{eq:Negative_definite_Hilbertian_metric}
D_{ij} = B_{ij}-\frac{1}{2}(B_{ii}+B_{jj}), 
\end{equation}
using the matrix $\mathbf{B}$ from proposition \ref{prop:infinite_divisibility_negative_definiteness}. There exists a Hilbert space $\mathcal{H}$ and a mapping $\phi$ such that
\begin{equation}\label{eq:Hilbertian_metric}
D_{ij} = \Vert\phi(i) - \phi(j) \Vert_{\mathcal{H}}^2.
\end{equation}
Moreover, notice that if $A$ is positive definite $-A$ is negative definite  and $\exp{A_{ij}}$ is infinitely divisible. In a similar way, we can construct a matrix,
\begin{equation}\label{eq:Positive_definite_Hilbertian_metric}
D_{ij} = -A_{ij}+\frac{1}{2}(A_{ii}+A_{jj}), 
\end{equation}
with the same property \eqref{eq:Hilbertian_metric}. This relation between \eqref{eq:Negative_definite_Hilbertian_metric} and \eqref{eq:Positive_definite_Hilbertian_metric} suggests a normalization of infinitely divisible matrices with non-zero diagonal elements that can be formalized in the following theorem.

\begin{figure}
\centering
\includegraphics[width =6cm]{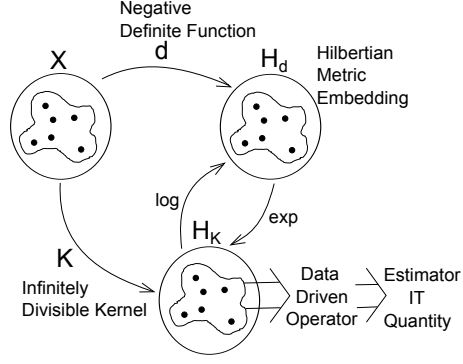}
\caption{Spaces involved in the infinitely divisible matrix framework}\label{fig:spaces} 
\end{figure}

\begin{thm}\label{thm:normalization_theorem} Let $\mathcal{X}$ be a nonempty set, and let $d_1$ and $d_2$ be two metrics on it, such that for any set $\{x_i\}_{i=1}^n$, $ \sum\limits_{i,j = 1}^n\alpha_i\alpha_j d_{\ell}^2(x_i,x_j)  \leq 0$, for any
$\bm{\alpha} \in \mathbb{R}^{n}$, and $\sum_{i=1}^n\alpha_i = 0$, is true for $\ell=1,2$. Consider the matrices $A_{ij}^{(\ell)} = \exp{-d_{\ell}^2(x_i,x_j)}$ and their normalizations $\hat{A}^{(\ell)}$, defined as:
\begin{equation}\label{eq:infinitely_divisible_normalization}
\hat{A}_{ij}^{(\ell)} = \frac{A_{ij}^{(\ell)}}{\sqrt{A_{ii}^{(\ell)}}\sqrt{A_{jj}^ {(\ell)}}}.
\end{equation}
Then, if $\hat{A}^{(1)} = \hat{A}^{(2)}$ for any finite set $\{x_i\}_{i=1}^n \subseteq \mathcal{X}$, there exist isometrically isomorphic Hilbert spaces $\mathcal{H}_1$ and $\mathcal{H}_2$, that contain the Hilbert space embeddings of the metric spaces $(\mathcal{X},d_{\ell})$, $\ell=1,2$. Moreover, $\hat{A}^{(\ell)}$ are infinitely divisible. 
\end{thm}  
Figure \ref{fig:spaces} summarizes the relation between spaces that are considered in the proposed framework. The object space $\mathcal{X}$ can be directly mapped into $\mathcal{H}_{\kappa}$ using an infinitely divisible kernel $\kappa$, or it can be mapped to a Hilbert space $\mathcal{H}_d$, if a negative definite function $d$, is employed as the distance function. The spaces $\mathcal{H}_\kappa$ and $\mathcal{H}_d$ are related by the $\log$ and $\exp$ functions. 

\section{Application to Metric Learning}
\subsection{Adaptation Using the Matrix-Based Entropy}
 By definition, the matrix entropy functional \eqref{eq:renyi_matrix_entropy} fall into the family of matrix functions know as spectral functions. These functions only depend on the eigenvalues of matrix and therefore their name \cite{SFriedland81}. 
Using theorem (1.1) from \cite{ALewis96a} it is straightforward to obtain the derivative of \eqref{eq:renyi_matrix_entropy} at $A$ as
\begin{equation}\label{eq:derivative_matrix_entropy}
\nabla S_{\alpha}(A) = \frac{\alpha}{(1-\alpha)\mathrm{tr}(A^{\alpha})}U \Lambda^{\alpha-1} U^*,
\end{equation}
where $A = U \Lambda U^*$. It is important to note that this decomposition can be used to our advantage. Instead of computing the full set of eigenvectors and eigenvalues of $A$, we can approximate the gradient of $S_\alpha$ by using only a few leading eigenvalues. It is easy to see that this approximation will be optimal in the Frobenius norm $\Vert \mathbf{X} \Vert_{\textrm{Fro}} = \sqrt{\mathrm{tr}(\mathbf{X}^{\mathrm{*}}\mathbf{X})}$.
\subsection{Metric Learning Using Conditional Entropy}
\begin{figure}[t]
\centering
\subfigure[Classification error UCI data]{\includegraphics[height = 6cm]{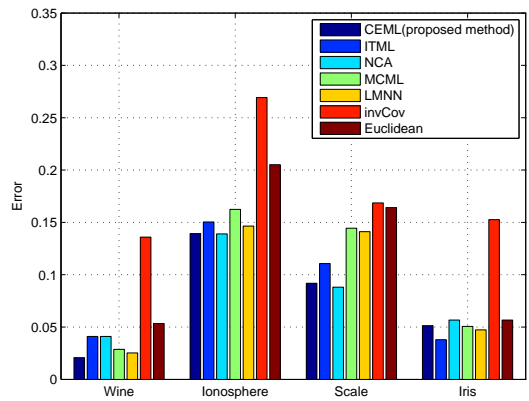}\label{fig:UCI_classification_error}}
\subfigure[Projected faces UMist dataset and resulting Gram matrix $\sigma = \sqrt{2}$]{\includegraphics[height = 6cm]{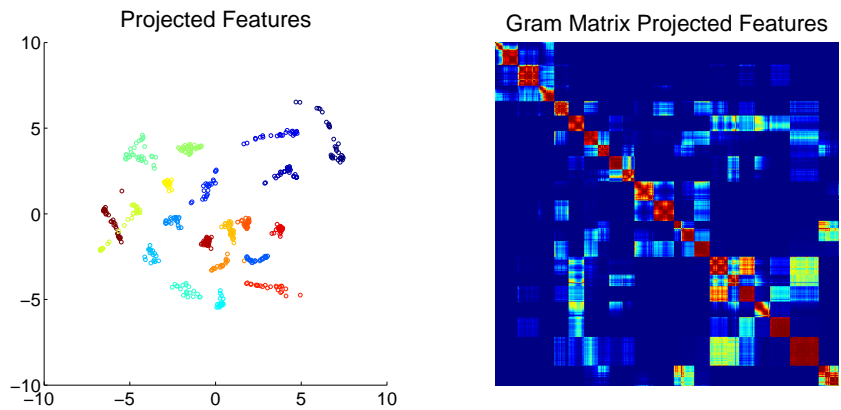}\label{fig:projected_faces}}
\caption{Results for the Metric learning application}\label{fig:results_metric_learning}
\end{figure}

\begin{figure}[h]
\centering
\includegraphics[height = 5cm]{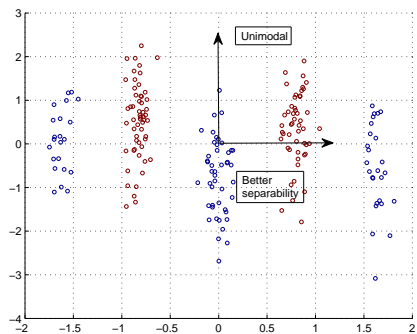}
\caption{Artificial data to illustrate the role of the entropy order}\label{fig:alpha_choice_data}
\end{figure}

Here, we apply the proposed matrix framework to the problem of supervised metric learning. This problem can be formulated as follows. Given a set of points $\{(\mathbf{x}_i,l_i)\}_{i=1}^n$, we seek a positive semidefinite matrix $\mathbf{AA}^{\mathrm{T}}$, that parametrizes a Mahalanobis distance between samples $\mathbf{x},\mathbf{x}^{\prime} \in \mathbb{R}^d$ as $d(\mathbf{x},\mathbf{x}^{\prime})=(\mathbf{x}-\mathbf{x}^{\prime})^{\mathrm{T}}\mathbf{AA}^{\mathrm{T}}(\mathbf{x}-\mathbf{x}')$. Our goal is to find parametrization matrix $\mathbf{A}$ such that the conditional entropy of the labels $l_i$ given the projected samples $\mathbf{y}_i = \mathbf{A}^{\mathrm{T}}\mathbf{x}_i$ with $\mathbf{y}_i \in \mathbb{R}^p$ and $p \ll d$, is minimized. This can be posed as the following optimization problem:
\begin{equation}\label{eq:metric_learning_optimization}
\begin{split}
\underset{\mathbf{A} \in \mathbb{R}^{d \times p}}{\textrm{minimize}} & \:\:S_{\alpha}(L \vert Y)\\
\textrm{subject to} &\:\: \begin{array}{l}
 \mathbf{A}^{\mathrm{T}}\mathbf{x}_i = \mathbf{y}_i,\:\textrm{for}\: i=1,\dots,n;\\
\mathrm{tr}(\mathbf{A}^{\mathrm{T}}\mathbf{A}) = p,
\end{array}
\end{split}
\end{equation}
where the trace constraint prevents the solution from growing unbounded. We can translate this problem to our matrix-based framework  in the following way. Let $\mathbf{K}$ be the matrix representing the projected samples \[K_{ij}  = \frac{1}{n}\exp{\left(-\frac{(\mathbf{x}_i-\mathbf{x}_j)^{\mathrm{T}}\mathbf{AA}^{\mathrm{T}}(\mathbf{x}_i-\mathbf{x}_j)}{2\sigma^2}\right)},\]
and $\mathbf{L}$ be the matrix of class co-occurrences where $L_{ij} = \frac{1}{n}$ if $l_i=l_j$ and zero otherwise. The conditional entropy can be computed as $S_{\alpha}(L|Y) = S_{\alpha}\left(n\mathbf{K} \circ \mathbf{L}\right) - S_{\alpha}(\mathbf{K})$, and its  gradient at $\mathbf{A}$, which can be derived based on \eqref{eq:spectral_function_derivative}, is given by:
\begin{equation}\label{eq:metric_learning_gradient}
\mathbf{X}^{\mathrm{T}}(\mathbf{P}-\mathrm{diag}(\mathbf{P}\mathbf{1}))\mathbf{X}\mathbf{A}
\end{equation}
where
\begin{equation}
\mathbf{P} = \left(n\mathbf{L}\circ\nabla S_{\alpha}\left(n\mathbf{K} \circ \mathbf{L}\right)-\nabla S_{\alpha}(\mathbf{K})\right)\circ \mathbf{K}
\end{equation}
Finally, we can use \eqref{eq:metric_learning_gradient} to search for $\mathbf{A}$ iteratively.\\
\textbf{UCI Data:} To evaluate the results we use the same experimental setup proposed in \cite{JDavis07}, we compares 5 different approaches to supervised metric learning based on the classification error obtained from  two-fold cross-validation using a $4$-nearest neighbor classifier. The reported errors are averages errors from 10 runs on the two folds for each algorithm; in our case the parameters are $p = 3$, $\alpha = 1.01$ and $\sigma = \sqrt{3}$. The feature vectors were centered and scaled to have unit variance. Figure \ref{fig:UCI_classification_error} shows the results of the proposed approach conditional entropy metric learning (CEML),  information theoretic metric learning (ITML) proposed in \cite{JDavis07}, neighborhood component analysis (NCA) from \cite{JGoldberger04}, the maximally collapsing metric learning (MCML) method from \cite{AGloberson05}, the large margin nearest neighbor (LMNN)  method found in \cite{KWeinberger05}, and, as a baseline, the the inverse covariance and Euclidean distances. The results for the \emph{Soybean} dataset are not reported since there is more than one possible data set in the UCI repository under that name.
The errors obtained by the metric learning algorithm using the proposed matrix-based entropy framework are consistently among the best performing methods included in the comparison.\\ 
\textbf{Choice of order $\alpha$:} Even though the choice of the entropy order above appears to be arbitrary, there is a motivation in choosing $\alpha$ close to $1$. The reason is that the higher the entropy order, the more prone the algorithm is to find unimodal solutions. This can be advantageous if prior knowledge or strong assumptions on the class distributions are taken into consideration. In our experiments, we opted for lower entropy order and give the algorithm more flexibility in finding a good solution. To experimentally show this phenomena, we generated a two-dimensional dataset containing points from two classes. In one direction the classes are very well separated but the distribution has multiple modalities. On the orthogonal direction, the classes are not fully separable, but their distributions are unimodal. Figure \ref{fig:alpha_choice_data} shows a sample with points drawn from both classes, as we can see projecting the data onto the horizontal axis provides better separability at the cost of a more complex decision boundary.
We run our metric learning algorithm 60 times for different values of $\alpha$ and recorded the direction of the resulting one-dimensional feature extractor. Table \ref{tab:alpha_choice} shows the number of times a particular direction was picked by our algorithm for different entropy orders. It can be seen that for larger values of $\alpha$, the algorithm selected the vertical direction more often.
\begin{table}[H]
\centering
\begin{tabular}{c||c|c|c|c}
$\alpha$ & $1.01$ & $1.3$ & $2$ & $5$  \\ \hline \hline 
Horizontal & $58$ & $35$  &  $1$ & $1$ \\
Vertical & $2$ & $25$ & $59$ & $59$ \\
\end{tabular}
\caption{Occurrence of horizontal and vertical solutions versus the entropy order}\label{tab:alpha_choice}
\end{table}
\textbf{UMist Faces:} We also run the algorithm on the UMist dataset; This data set consists of Grayscale faces (8 bit [0-255]) of 20 different people.
The total number of images is 575 and the size of each image is 112x92 pixels for a total of $10304$ dimensions. Pixel values were normalized by dividing by $255$ and removing the mean. Figure \ref{fig:projected_faces} shows the images projected into $\mathbb{R}^2$. It is remarkable how a linear projection can separate the faces, and it can also be seen from the Gram matrix that it tries to approximate the co-occurrence matrix $\mathbf{L}$.

\section{Conclusions}
In this paper, we presented a data-driven framework for information theoretic learning based on infinitely divisible matrices. We define estimators of entropy-like quantities that can be computed from the Gram matrices obtained by evaluating infinitely divisible kernels on pairs of samples. The proposed quantities do not assume that the density of the data has been estimated, this can be advantageous in many scenarios where even defining a density is not feasible. We discuss some  key properties of the proposed quantities and show how they can be applied to define useful analogues to quantities such as conditional entropy. Based on the proposed framework, we introduce a supervised metric learning algorithm with results that are competitive with the state of the art. Nevertheless, we believe that many interesting formulations to learning problems based on the proposed framework are yet to be found. It is also important to highlight that the connection between the RKHS provided by the infinitely divisible kernel, and the Hilbertian metrics associated with the negative definite functions, opens an interesting avenue to investigate formulations of information theoretic learning algorithms on both spaces, and the implications of choosing one or the other. 

%
\bibliographystyle{IEEEtran}
\bibliography{biblio}
\appendix
\section{Additional results and proofs}
To prove \eqref{eq:hadamard_conditional_entropy_inequality}, we need to introduce the concept of majorization and some results pertaining the ordering that arises from this definition. The proposition is replicated in this appendix for the sake of self containment.
\begin{defi}\emph{(Majorization): }Let $p$ and $q$ be two nonnegative vectors in $\mathbb{R}^n$ such that $\sum_{i=1}^n p_i = \sum_{i=1}^n q_i < \infty$. We say $p \preccurlyeq q$, $q$ majorizes $p$, if their respective ordered sequences $p_{[1]} \geq p_{[2]} \geq \dots \geq p_{[n]}$  and $q_{[1]} \geq q_{[2]} \geq \dots \geq q_{[n]}$ denoted by $\{p_{[i]}\}_{i=1}^n$ and $\{p_{[i]}\}_{i=1}^n$, satisfy: 
\begin{equation}
\sum_{i=1}^k p_{[i]}  \leq \sum_{i=1}^k q_{[i]}\:\: \textrm{for} \:\: k =1,\dots, n
\end{equation}
\end{defi}
It can be shown that if $p \preccurlyeq q$ then $p = Aq$ for some doubly stochastic matrix $A$ \cite{RBhatia}. It is also easy to verify that if $p \preccurlyeq q$ and $p \preccurlyeq h$ then $p \preccurlyeq tq +  (1-t)h$ for $t \in [0,1]$. The majorization order is important because it can be associated with the definition of \emph{Schur-concave (convex)} functions. A real valued function $f$ on $\mathbb{R}^n$ is called Schur-convex if $p \preccurlyeq q$ implies $f(p) \leq f(q)$ and Schur-concave if $f(q) \leq f(p)$. 
\begin{lmm}\label{lmm:Renyi_schur_concavity} The function $f_{\alpha}:\mathcal{S}^n \mapsto \mathbb{R}_{+}$ ($\mathcal{S}^n$ denotes the $n$ dimensional simplex), defined as,
\begin{equation}
f_{\alpha}(p) = \frac{1}{1-\alpha}\log_2{\sum\limits_{i=1}^n p_i^{\alpha}}, 
\end{equation}
is Schur-concave for $\alpha>0$.
\end{lmm}
Notice that, Schur-concavity (Schur-convexity) cannot be confused with concavity (convexity) of a function in the usual sense. Now, we are ready to state the inequality for Hadamard products. 
\begin{prop} Let $A$ and $B$ be two $n \times n$ positive definite matrices with trace $1$ with nonnegative entries, and $A_{ii} = \frac{1}{n}$ for $i=1,2,\dots,n$. Then, the following inequality holds:
 \begin{equation} 
 S_{\alpha}\left(\frac{A \circ B}{\mathrm{tr}(A \circ B)}\right) \geq S_{\alpha}(B),
\end{equation}
\begin{prf} In proving \eqref{eq:hadamard_conditional_entropy_inequality}, we will use the fact that $S_{\alpha}$ preserves the majorization order (inversely) of nonnegative sequences on the $n$-dimensional simplex.
First look at the identity
\[
x^{\mathrm{T}}(A \circ B)x = \mathrm{tr}{\left(A D_x B D_x\right)} = \frac{1}{n}
\]
In particular, if $\{x_i\}_{i=1}^n$ is an orthonormal basis for $\mathbb{R}^n$,
$
\mathrm{tr}{\left(A \circ B \right)} = \sum\limits_{i=1}^nx_i^{\mathrm{T}}(A \circ B)x_i 
$. If we let  $\{x_i\}_{i=1}^n$ be the eigenvectors of $A \circ B$ ordered according to their respective eigenvalues in decreasing order, then,
\begin{eqnarray}
\nonumber \sum\limits_{i=1}^k x_i^{\mathrm{T}}(A \circ B)x_i =  \sum\limits_{i=1}^k \mathrm{tr}\left(A D_{x_i} B D_{x_i}\right) & \leq &  \frac{1}{n}\sum\limits_{i=1}^k \mathrm{tr}\left(\mathbf{1}\mathbf{1}^{\mathrm{T}} D_{x_i} B D_{x_i}\right)\\
\label{eq:majorization_conditional_inequality} & = & \frac{1}{n}\sum\limits_{i=1}^k x_i^{\mathrm{T}} B x_i  \leq  \frac{1}{n}\sum\limits_{i=1}^k y_i^{\mathrm{T}} B y_i,
\end{eqnarray}
where $k=1,\dots,n$ and $\{y_i\}_{i=1}^n$ are the eigenvectors of $B$ ordered according to their respective eigenvalues in decreasing order. The inequality \eqref{eq:majorization_conditional_inequality} is equivalent  to say that $n \lambda(A \circ B) \preccurlyeq \lambda(B)$, that is, the sequence of eigenvalues of $(A \circ B)/\mathrm{tr}{(A \circ B)}$ is majorized by the sequence of eigenvalues of $B$, which implies \eqref{eq:hadamard_conditional_entropy_inequality} by Lemma \ref{lmm:Renyi_schur_concavity}.
\end{prf}
\end{prop}

A beautiful observation from Theorem \ref{thm:normalization_theorem} is that, according to equation \eqref{eq:conditonal_entropy_gap}, the proposed normalization procedure for infinitely divisible matrices can be thought of as finding the maximum entropy matrix among all matrices for which the Hilbert space embeddings are isometrically isomorphic. 
\subsection{Derivatives of Spectral Functions}
Let $H_n$ denote the vector space of real Hermitian matrices of size $n \times n$ endowed with inner product $\langle \mathbf{X}, \mathbf{Y}\rangle = \mathrm{tr}\mathbf{X}\mathbf{Y}$; and let $U_n$ denote the set of $n \times n$ unitary matrices. A real valued function $f$ defined on a subset of $H_n$ is unitarily invariant if $f(\mathbf{U}\mathbf{X}\mathbf{U}^*) = f(\mathbf{X})$ for any $\mathbf{U} \in U_n$. Associated with each spectral function $f$ there is a symmetric function $F$ on $\mathbb{R}^n$. By symmetric we mean that $F(\mathbf{x}) = F(\mathbf{P}\mathbf{x})$ for any $n \times n$ permutation matrix $\mathbf{P}$. Let $\lambda(\mathbf{X})$ denote the vector of ordered eigenvalues of $\mathbf{X}$; then, a spectral function $f(\mathbf{X})$ is of the form $F(\lambda(\mathbf{X}))$ for $F$ a symmetric. We are interested in the differentiation of the composition $(F\circ\lambda)(\cdot) = F(\lambda(\cdot))$ at $\mathbf{X}$\footnote{In here, $\circ$ denotes composition rather than Hadamard product}. The following result \cite{ALewis96a} allows us to differentiate a spectral function $f$ at $\mathbf{X}$
\begin{thm}\label{thm:spectral_function_derivative}
Let the set $\Omega \subset \mathbf{R}^n$ be open and symmetric, that is, for any $\mathbf{x} \in \Omega$ and any $n\times n$ permutation matrix $\mathbf{P}$, $\mathbf{P}\mathbf{x} \in \Omega$. Suppose that $F$ is symmetric, Then, the spectral function $F(\lambda(\cdot))$ is differentiable at a matrix $\mathbf{X}$ if and only if $F$ is differentiable at the vector $\lambda(X)$. In this case, the gradient of $F\circ\lambda$ at $\mathbf{X}$ is
\begin{equation}\label{eq:spectral_function_derivative}
\nabla(F\circ \lambda)(\mathbf{X}) = \mathbf{U}\mathrm{diag(\nabla F(\lambda(\mathbf{X})))}\mathbf{U}^* ,
\end{equation}
for any unitary matrix satisfying $\mathbf{X} = \mathbf{U}\mathrm{diag(\lambda(\mathbf{X}))}\mathbf{U}^{*}$.
\end{thm}

\end{document}